\documentclass[10pt,twocolumn,letterpaper]{article}

\usepackage{cvpr}              

\usepackage{graphicx}
\usepackage{amsmath}
\usepackage{amssymb}
\usepackage{booktabs}
\usepackage{multirow}
\usepackage{adjustbox}
\usepackage{mathtools}
\usepackage{subcaption}

\usepackage[pagebackref,breaklinks,colorlinks]{hyperref}

\usepackage[capitalize]{cleveref}
\crefname{section}{Sec.}{Secs.}
\Crefname{section}{Section}{Sections}
\Crefname{table}{Table}{Tables}
\crefname{table}{Tab.}{Tabs.}

\def\cvprPaperID{2103} 
\def\confName{CVPR}
\def\confYear{2023}

\begin{document}

\title{Meta Compositional Referring Expression Segmentation}

\author{Li Xu$^1$, Mark He Huang$^{1,3}$, Xindi Shang$^2$, Zehuan Yuan$^2$, Ying Sun$^3$, 
Jun Liu $^{1}$\thanks{Corresponding Author}\\
$^1$Singapore University of Technology and Design, Singapore \\ 
$^2$ByteDance\\
$^3$Institute for Infocomm Research (I$^2$R) \& Centre for Frontier AI Research (CFAR), A*STAR, Singapore \\
{\tt\small \{li\_xu, he\_huang\}@mymail.sutd.edu.sg}\\
{\tt\small \{shangxindi, yuanzehuan\}@bytedance.com}\\
{\tt\small suny@i2r.a-star.edu.sg, jun\_liu@sutd.edu.sg}
}
\maketitle
\begin{abstract}
   Referring expression segmentation aims to segment an object described by a language expression from an image.
   Despite the recent progress on this task, existing models tackling this task may not be able to fully capture semantics and visual representations of individual concepts, which limits their generalization capability, especially when handling \textbf{novel compositions of learned concepts}.
   In this work, through the lens of meta learning, we propose a Meta Compositional Referring Expression Segmentation (\textbf{MCRES}) framework to enhance model compositional generalization performance.
   Specifically, to handle various levels of novel compositions, our framework first uses training data to construct a virtual training set and multiple virtual testing sets, where data samples in each virtual testing set contain a level of novel compositions w.r.t. the virtual training set.
   Then, following a novel meta optimization scheme to optimize the model to obtain good testing performance on the virtual testing sets after training on the virtual training set,
   our framework can effectively drive the model to better capture semantics and visual representations of individual concepts, and thus obtain robust generalization performance even when handling novel compositions.
   Extensive experiments on three benchmark datasets demonstrate the effectiveness of our framework.
\end{abstract}

\section{Introduction}\label{sec:intro}

\begin{figure}[t] \label{fig:example}
   \centering
   \begin{subfigure}{\linewidth}
      \centering
      \includegraphics[width=0.85\linewidth]{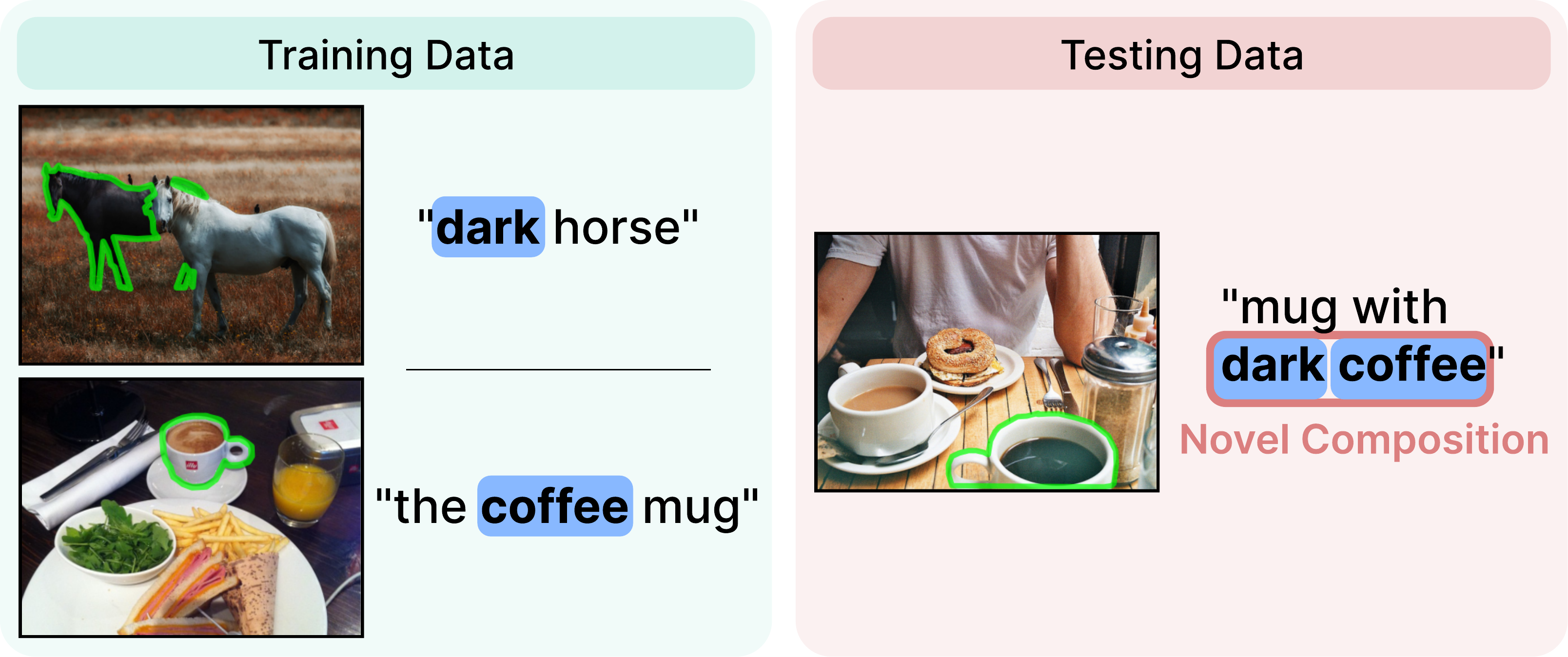}
      \caption{ }
      \label{fig:example-a}
   \end{subfigure}
   \begin{subfigure}{\linewidth}
      \centering
      \includegraphics[width=0.82\linewidth]{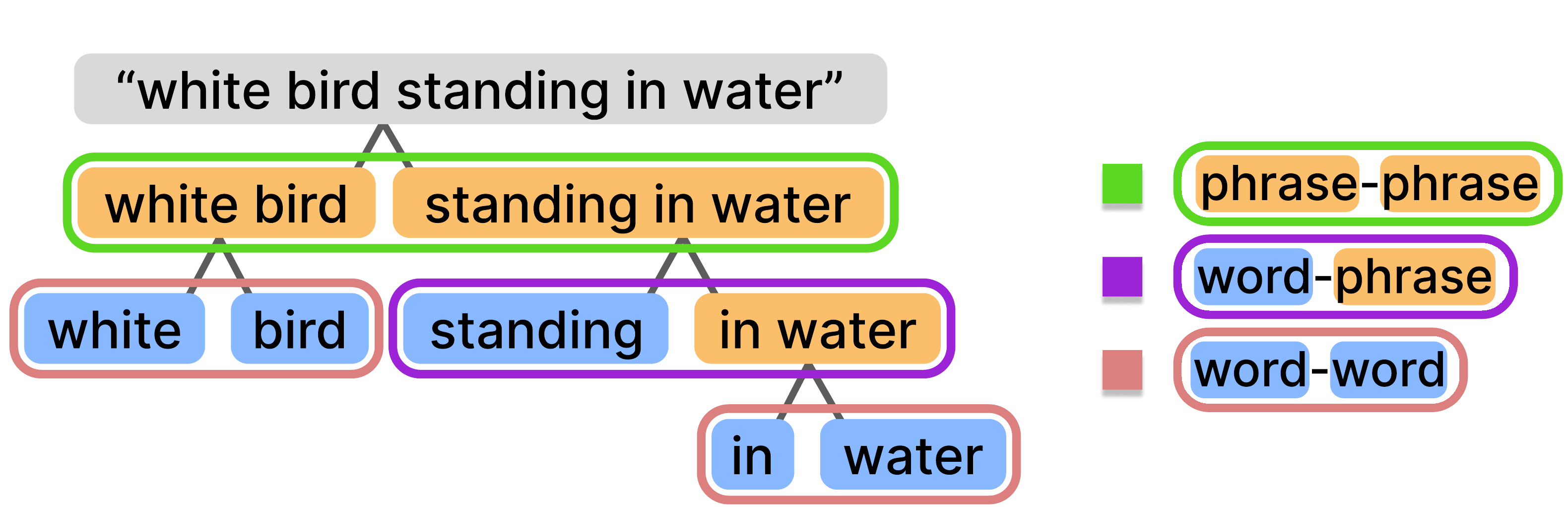}
      \caption{ }
      \label{fig:example-b}
   \end{subfigure}
   \caption{\textbf{Illustration of novel compositions and various levels of compositions.}
     (a) An example of the testing sample containing the novel composition of ``dark coffee" in RefCOCO dataset \cite{yu2016modeling}.
    Such a novel composition itself does not exist in the training data, but its individual components (i.e., ``dark'' and ``coffee'') exist in the training data.
    (b) An example of various levels of compositions in an expression.
    We perform constituency parsing of the expression using AllenNLP \cite{gardner2018allennlp}. Based on the obtained parsing tree as shown above, we can then obtain various levels of compositions (e.g., word-word level, word-phrase level) in this expression.}
    \vspace{-0.6cm}
\end{figure}

Referring expression segmentation (RES) \cite{hu2016segmentation, yu2018mattnet, ye2019cross} aims to segment a visual entity in an image given a linguistic expression.
This task has been receiving increasing attention in recent years \cite{wang2022cris, yang2022lavt, kim2022restr,ding2021vision},
as it can play an important role in various applications, such as language-based human-robot interaction and interactive image editing.
However, despite the recent progress on tackling this task \cite{ding2021vision, wang2022cris, zhu2022seqtr},
existing methods may struggle with handling the testing samples, of which the expressions contain novel compositions of learned concepts.
Here a novel composition means that the composition itself does not exist in the training data, but its individual components (e.g., words, phrases) exist in the training data, as shown in Fig. \ref{fig:example-a}.

We observe that testing samples containing such novel compositions of learned concepts widely exist in RES datasets \cite{yu2016modeling, nagaraja2016modeling, mao2016generation}. 
However, existing RES models may not be able to well handle novel compositions during testing.
Here we test the generalization capability of multiple state-of-the-art models \cite{yang2022lavt, zhu2022seqtr, luo2020multi} in terms of handling novel compositions, as shown in Table \ref{Tab:ablation_1}.
Specifically, we first split each testing set of the RefCOCO dataset. 
In each testing set, one split subset includes the data samples, in which all the contained compositions are seen in the RefCOCO training set.
While another subset includes the data samples containing novel compositions, of which the individual components (e.g., words, phrases) exist in the RefCOCO training set but the composition itself is unseen in the training set, i.e., containing novel compositions of learned concepts.
Then we evaluate the models \cite{yang2022lavt, zhu2022seqtr, luo2020multi} on the two subsets in each testing set, and find that for each model, its testing performance on the subset containing novel compositions drops obviously compared to the performance on the other subset.
For these models, the performance gap between the two subsets can reach $14\%-17\%$ measured by the metric of overall IoU.
Such a clear performance gap indicates that existing models struggle with generalizing to novel compositions of learned concepts.
This might due to that the model does not effectively capture the semantics and visual representations of individual concepts (e.g., ``dark", ``coffee" in Fig. \ref{fig:example-a}) during training. Then the trained model may fail to recognize a novel composition (e.g., ``dark coffee") at testing time, which though is composed of learned concepts.

Thus to handle this issue,
we aim to train the model to effectively capture the semantics and visual representations of individual concepts during training.
Despite the conceptual simplicity, how to guide the model's learning behavior towards this goal is a challenging problem.
Here from the perspective of meta learning,
we propose a \emph{Meta Compositional Referring Expression Segmentation (\textbf{MCRES})} framework, to effectively handle such a challenging problem by only changing the model training scheme.

Meta learning proposes to perform \emph{virtual testing} during model training for better performance \cite{finn2017model, nichol2018first}.
Inspired by this, to improve the generalization capability of RES models, 
our MCRES framework incorporates a \emph{meta optimization} scheme that consists of three steps: \emph{virtual training}, \emph{virtual testing} and \emph{meta update}.
Specifically, we first split the training set to construct a virtual training set for virtual training, and a virtual testing set for virtual testing.
The data samples in the virtual testing set contain novel compositions w.r.t. the virtual training set.
For example, if the expressions of data samples in the virtual training set contain both words ``dark" and ``coffee" but do not contain their composition (i.e., ``dark coffee"),
the virtual testing set can include this novel composition correspondingly.

Based on the constructed virtual training set and virtual testing set, we first train the model using the virtual training set, and then evaluate the trained model on the virtual testing set.
During virtual training, the model may learn the compositions of individual concepts as a whole without truly understanding the semantics and visual representations of individual concepts,
which though can still improve model training performance.
For example, if there are many training samples containing the composition of ``yellow banana" in the virtual training set,
the model can superficially correlate ``banana" with ``yellow" and learn this composition as a whole, 
since using such spurious correlations can facilitate the model learning \cite{arjovsky2019invariant, ye2021adversarial, geirhos2020shortcut}.
However, learning the compositions as a whole over the virtual training set may not improve model performance much on the virtual testing set in virtual testing,
since the virtual testing set contains novel compositions w.r.t. the virtual training set.
Thus to achieve good testing performance on such a virtual testing set, the model needs to effectively capture semantics and visual representations of individual concepts during virtual training.
In this way, the model testing performance on the virtual testing set serves as a generalization feedback to the model virtual training process.

Thus after the virtual training and virtual testing, we can further update the model to obtain better testing performance on the virtual testing set (i.e., meta update), so as to drive the model training on the virtual training set towards the direction of learning to capture semantics and visual representations of individual concepts,
i.e., learning to learn.
In this manner, our framework is able to optimize the model for robust generalization performance, even tackling the challenging testing samples with novel compositions.

Moreover, given that expressions can often be hierarchically decomposed,
there can exist various levels of novel compositions.
Specifically, to identify meaningful compositions in an expression, we can parse an expression into a tree structure based on the constituency parsing tool \cite{gardner2018allennlp} as shown in Fig.\ref{fig:example-b}.
In such a parsing tree, under the same parent node, 
each pair of child nodes (e.g., ``white" and ``bird") are closely semantically related, and thus can form a meaningful composition.
Since each child node can be a word or a phrase as in Fig. \ref{fig:example-b}, there can naturally exist the following three levels of novel compositions:
word-word level (e.g., ``white" and ``bird"), word-phrase level (e.g., ``standing" and ``in water") and phrase-phrase level (e.g., ``white bird" and ``standing in water"),
which correspond to different levels of comprehension complexity.
To better handle such a range of novel compositions,
we construct multiple virtual testing sets in our framework, where each virtual testing set is constructed to handle one level of novel compositions.

Our framework only changes the model training scheme without the need to change the model structure.
Thus our framework is general, and can be conveniently applied on various RES models.
We test our framework on multiple models, and obtain consistent performance improvement.

The contributions of our work are threefold:
1) We propose a novel framework (MCRES)
to effectively improve generalization performance of RES models, especially when handing novel compositions of learned concepts.
2) Via constructing a virtual training set and multiple virtual testing sets w.r.t. various levels of novel compositions,
our framework can train the model to well handle various levels of novel compositions.
3) When applied on various models on three RES benchmarks \cite{yu2016modeling, nagaraja2016modeling},
our framework achieves consistent performance improvement.

\section{Related Works}

\textbf{Referring Expression Segmentation (RES).} RES \cite{hu2016segmentation} aims to segment a target object from an image based on an expression.
Early works \cite{li2018referring, liu2017recurrent, yu2016modeling, nagaraja2016modeling} employed convolutional and recurrent networks to extract visual and linguistic features respectively,
and then fused the extracted features to predict the segmentation mask.
Recently, a series of works \cite{ding2021vision, kim2022restr, yang2022lavt} employed vision and language transformers to boost performance.
Besides, with the development of large-scale pretrained models, Wang et al. \cite{wang2022cris} proposed to leverage CLIP \cite{radford2021clip} to improve cross-modal matching.

Some methods have been explored to help model better understand learned concepts in RES.
Yu et al. \cite{yu2018mattnet} proposed a modular network that uses different modules to process different types of information in the given expression.
Yang et al. \cite{yang2021bottom} designed a reasoning module to help align the language concepts with visual regions.
Different from all the above-mentioned works,  we propose an MCRES framework to drive RES models to better capture semantics and visual representations of individual concepts.
Such a framework only changes the model training scheme, and thus can be flexibly applied on various models to improve their generalization performance.

\textbf{Meta Learning.} Meta learning, i.e., the paradigm of learning to learn,
has emerged to mainly tackle the few-shot learning problem \cite{snell2017prototypical, finn2017model, rajeswaran2019meta, nichol2018first}.
MAML \cite{finn2017model} and its following works \cite{nichol2018first, rajeswaran2019meta} aim to learn a good initialization of network parameters, to achieve fast test-time update to adapt to new few-shot learning tasks.
More recently, meta learning has also been explored in other areas \cite{li2018learning, guo2020learning, huang2021metasets,foo2022era, xu2022meta} to enhance model generalization performance without the need of test-time update. 
Inspired by these works, we leverage a meta learning-based framework to improve generalization performance of RES models especially when handling novel compositions. 

\section{Method}

\begin{figure*}[t]\label{fig:method}
   \centering
   \begin{subfigure}{0.50\linewidth}
      \centering
      \includegraphics[width=0.88\linewidth]{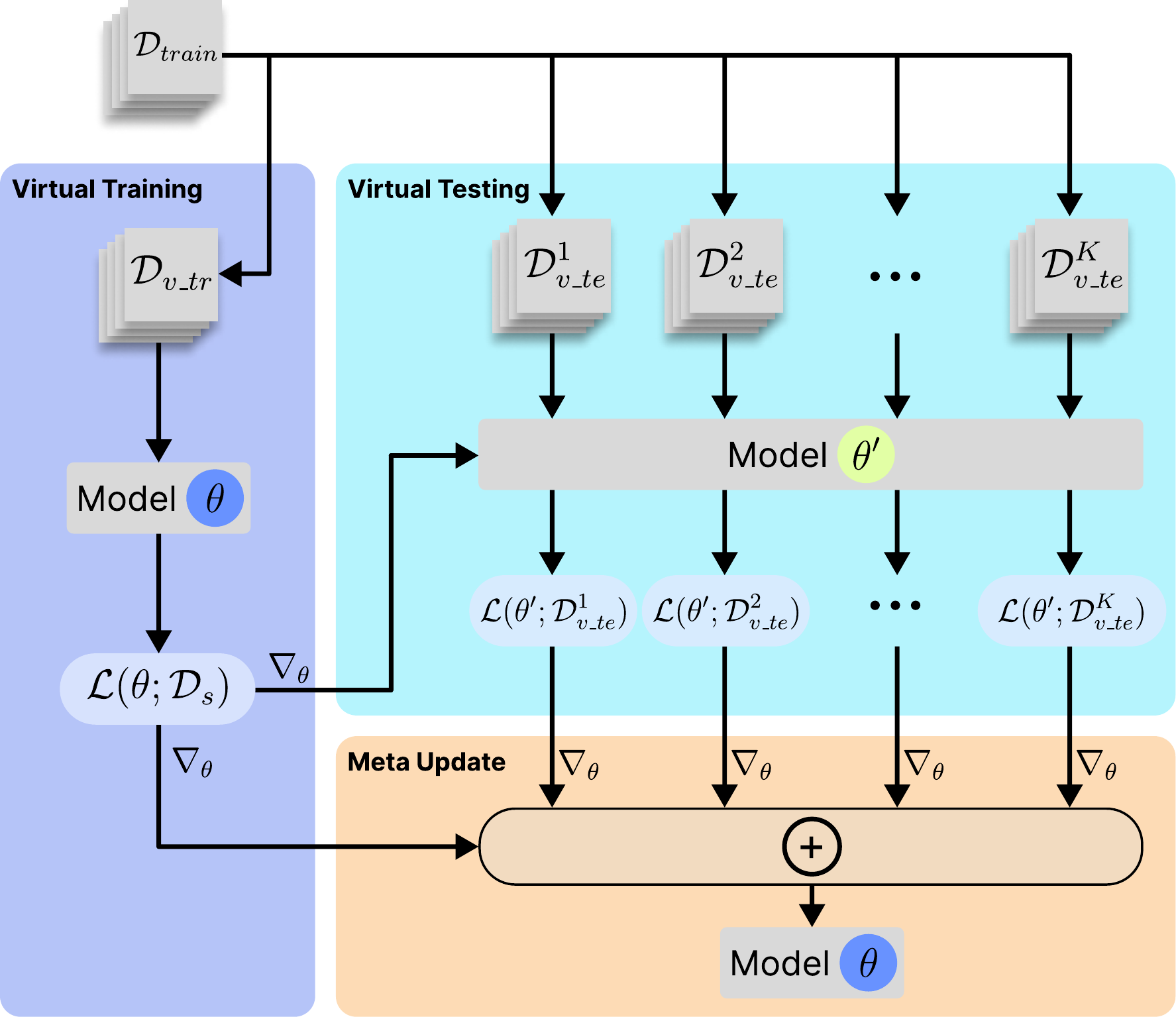}
      \caption{ }
      \label{fig:method-a}
   \end{subfigure}
   \hfill
   \begin{subfigure}{0.48\linewidth}
      \centering
      \includegraphics[width=0.83\linewidth]{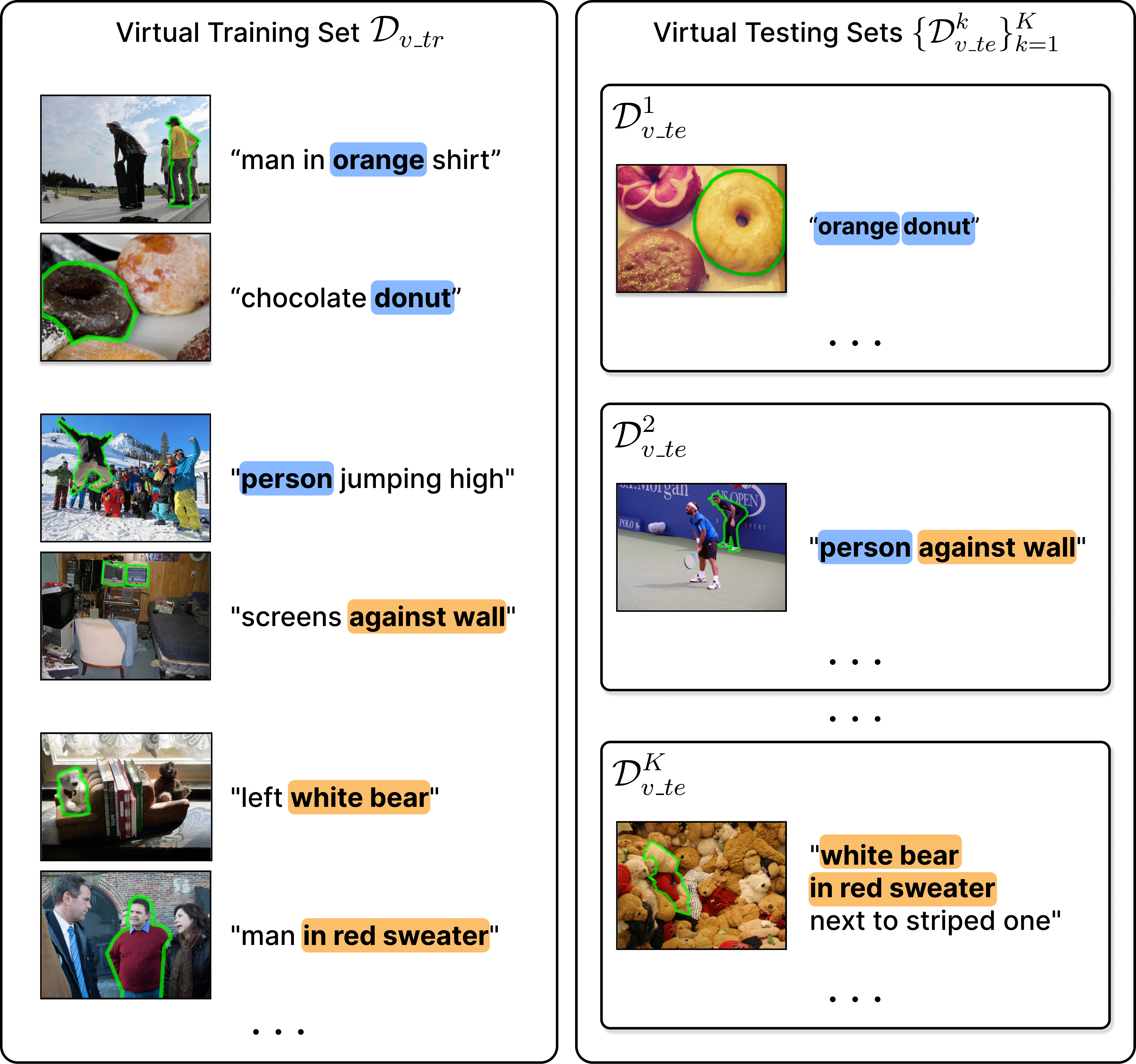}
      \caption{ }
      \label{fig:method-b}
   \end{subfigure}
   \vspace{-0.3cm}
   \caption{\textbf{Framework overview.}
   Fig. (a) illustrates our framework pipeline.
   First, we train the model using the virtual training set ($\mathcal{D}_{v\_tr}$), and then obtain the updated model.
   We then test the model with updated parameters ($\theta^{\prime}$) on multiple virtual testing sets ($\{{\mathcal{D}^{k}_{v\_te}}\}^K_{k=1}$).  
   According to the virtual testing losses,
   we perform meta update to optimize the model for better generalization capability.
   Fig. (b) shows that 
   we construct a virtual training set and multiple virtual testing sets to handle various levels of novel compositions.
   The expressions of data samples in each virtual testing set 
   contain a level of novel compositions w.r.t. the virtual training set. 
   }
   \label{fig:short}
   \vspace{-0.4cm}
\end{figure*}

Existing RES models may fail to generalize to data samples containing novel compositions of learned concepts.
To handle this issue, we aim to encourage the model to better capture the semantics of individual concepts as well as recognize their visual correspondences during training, which however is a non-trivial problem.
In this paper, from the perspective of meta learning, we introduce MCRES framework optimizing RES models via a meta optimization scheme to improve their generalization performance.

Specifically, as shown in Fig. \ref{fig:short}, in our framework, we first split the original training set ($\mathcal{D}_{train}$) to build a virtual training set ($\mathcal{D}_{v\_tr}$) for virtual training, and a group of $K$ virtual testing sets ($\{{\mathcal{D}^{k}_{v\_te}}\}^K_{k=1}$) for virtual testing.
Each virtual testing set consists of data samples containing one level of novel compositions w.r.t. the virtual training set.
We first use the virtual training set to train the model (virtual training), and then perform model testing on the virtual testing sets (virtual testing).
Since the virtual testing sets contain novel compositions of individual concepts from the virtual training set,
if the model can still achieve robust testing performance on the virtual testing sets after training on the virtual training set, we posit that the trained model has learned to capture more semantics and visual representations of individual concepts during the virtual training.
Guided by this principle, we can optimize the model virtual testing performance,
to guide the model virtual training process towards learning more semantics and representations of individual concepts.
Below, we first introduce the pipeline of our framework, and then detail the virtual training set and virtual testing sets construction process.

\subsection{Framework} \label{sec:meta_optimization}

We first train the RES model using the virtual training set, i.e., virtual training.
Specifically, we denote the model parameters as $\theta$, the loss function (e.g., cross-entropy loss) for training the RES model as $\mathcal{L}$.
Thus we can calculate the model virtual training loss ($\mathcal{L}_{v\_tr}$) over the virtual training set ($\mathcal{D}_{v\_tr}$) as:
\begin{equation}
 \setlength{\abovedisplayskip}{6pt}
 \setlength{\belowdisplayskip}{6pt}
   \mathcal{L}_{v\_tr} (\theta) = \mathcal{L}(\theta;\mathcal{D}_{v\_tr})
\end{equation}
Based on this loss, we then update the model parameters ($\theta$) as follows:
\begin{equation}
 \setlength{\abovedisplayskip}{6pt}
 \setlength{\belowdisplayskip}{6pt}
   \theta^{\prime} = \theta - \alpha \nabla_{\theta} \mathcal{L}_{v\_tr} (\theta)
\end{equation}
where $\alpha$ denotes the learning rate.
It is worth noting that the model update at this step is \emph{virtual}, and the updated parameters $\theta^{\prime}$ only serve as intermediate parameters for the model evaluation at the following virtual testing step.

After the virtual training, we evaluate the generalization performance of the trained model 
on the virtual testing sets ($\{{\mathcal{D}^{k}_{v\_te}}\}^K_{k=1}$), i.e., virtual testing.
Specifically, on each virtual testing set $\mathcal{D}^{k}_{v\_te}$, we compute the model loss $\mathcal{L}^{k}_{v\_te}$ as:
\begin{equation}
 \setlength{\abovedisplayskip}{6pt}
 \setlength{\belowdisplayskip}{6pt}
   \mathcal{L}^{k}_{v\_te} (\theta^{\prime}) = \mathcal{L}(\theta^{\prime};\mathcal{D}^{k}_{v\_te})
\end{equation}
Such a loss measures how well the model generalizes to the virtual testing set after training on the virtual training set. 

Then we perform the meta update step. 
At this step, we wish to \emph{actually} update the model parameters ($\theta$), so that after the model training on the virtual training set, 
the trained model can generalize well
to the virtual testing sets, i.e., generalizing well to novel compositions.
To this end, we formulate the optimization objective as:
\begin{equation}\label{eq:optimization-objective}
   \setlength{\abovedisplayskip}{6pt}
   \setlength{\belowdisplayskip}{6pt}
   \begin{aligned}
        & \min_{\theta}\;\mathcal{L}_{v\_tr}(\theta) + \sum_{k=1}^K \mathcal{L}_{v\_te}^k(\theta^{\prime})                                                     \\
      = & \min_{\theta}\;\mathcal{L}_{v\_tr}(\theta) + \sum_{k=1}^K \mathcal{L}_{v\_te}^k \big (\theta - \alpha\nabla_\theta \mathcal{L}_{v\_tr}(\theta) \big)
   \end{aligned}
\end{equation}
where the first term indicates the model training performance on the virtual training set, and the second term represents the model testing performance on the virtual testing sets after training on the virtual training set.
Based on this objective, we then update the model parameters ($\theta$) as:
\begin{equation}\label{eq:meta-update}
   \setlength{\abovedisplayskip}{6pt}
   \setlength{\belowdisplayskip}{6pt}
   \begin{aligned}
      \theta \leftarrow \theta - \beta \nabla_{\theta} \Big(\mathcal{L}_{v\_tr}(\theta) + \sum_{k=1}^K \mathcal{L}_{v\_te}^k \big (\theta - \alpha\nabla_\theta \mathcal{L}_{v\_tr}(\theta) \big) \Big)
   \end{aligned}
\end{equation}
where $\beta$ is the learning rate for meta update.
Through the above optimization process,
the model is pushed to learn more semantics and visual representations of individual concepts during training.

During the above process, the model is first trained (updated) on the virtual training set.
At this step, 
the model may learn the compositions of individual concepts as a whole without effectively capturing the semantics and visual representations of individual concepts, which though can still improve its training performance on the virtual training set.
However, to achieve good testing performance on the virtual testing data
which contains novel compositions of individual concepts from the virtual training data,
the model is expected to avoid learning compositions as a whole and instead
capture more semantics and representations of individual concepts.
In this way, the second term of Eqn. \ref{eq:meta-update}, which includes the second-order gradients of $\theta$: $\nabla_\theta \mathcal{L}_{v\_te}^k \big (\theta - \alpha\nabla_\theta \mathcal{L}_{v\_tr}(\theta) \big)$, can be regarded as a generalization feedback that can guide the model
to capture more semantics and representations of individual concepts.

In our framework, the above three steps (i.e., virtual training, virtual testing and meta update) are performed iteratively until the model training converges.

\subsection{Sets Construction} \label{sec:splitting}

In our framework, to handle various levels of novel compositions, we split the original training set to construct a virtual training set and various virtual testing sets.
Each virtual testing set is expected to contain one level of novel compositions w.r.t. the virtual training set.
Below we first introduce the details of these sets, and then discuss the strategy for constructing the virtual testing sets. 

We randomly sample a subset of the training set ($\mathcal{D}_{train}$) as the virtual training set ($\mathcal{D}_{v\_tr}$), and the remaining training data will be used as the candidate samples to construct a group of ($K$) virtual testing sets ($\{{\mathcal{D}^{k}_{v\_te}}\}^K_{k=1}$).
Each virtual testing set is constructed to handle one level of novel compositions.
Here considering the hierarchical semantic structure of each expression, we target the following levels of novel compositions: word-word level, word-phrase level and phrase-phrase level.
A phrase means a group of words that serve as a grammatical unit in the expression (e.g., ``white bird" in Fig. \ref{fig:example-b}), which can be conveniently identified using off-the-shelf tools
(e.g., AllenNLP \cite{gardner2018allennlp}).

Specifically, to identify various levels of novel compositions in an expression, we first use the constituency parsing tool \cite{gardner2018allennlp} to parse an expression into a tree structure as shown in Fig. \ref{fig:example-b}.
In this tree, under the same parent node, 
each pair of child nodes 
are closely related, and thus can form a meaningful composition.
If a pair of nodes having the same parent node are both words, they form a word-word level composition (e.g., ``white" and ``bird").
Similarly, if the nodes with the same parent node are a word and a phrase, they form a word-phrase level composition (e.g., ``standing" and ``in water").
In a similar way, we can obtain phrase-phrase level compositions (e.g., ``white bird" and ``standing in water").
Thus there can exist three levels of novel compositions, which correspond to different levels of comprehension complexity.
Corresponding to these three levels of novel compositions, we will construct a total of three virtual testing sets ($K=3$).

\textbf{Virtual testing sets construction.}
To construct each virtual testing set, we need to select data samples containing the corresponding level of novel compositions w.r.t. the virtual training set, from the candidate samples.
A data sample means an expression paired with the corresponding image.
Here we design an efficient strategy, which can be leveraged to construct each virtual testing set.
Below, we take the process of constructing the virtual testing set for handling \emph{word-word} level novel compositions as an example, to introduce such a strategy.
Specifically, this virtual testing set should include all of those candidate samples that contain word-word level novel compositions w.r.t. the virtual training set.
To construct such a virtual testing set, our strategy proceeds as follows.

(i) To find the data samples, of which the expressions contain word-word level novel compositions w.r.t. the virtual training set, from the candidate samples,
we need to first obtain all the word-word level compositions in the virtual training set and in the candidate samples respectively.
Thus by parsing each expression into a parsing tree as shown in Fig. \ref{fig:example-b},
we can obtain all word-word level compositions (e.g., ``white" and ``bird", ``in" and ``water" in Fig. \ref{fig:example-b}) in each expression.
(ii) To identify word-word level novel compositions, we first select the word-word level compositions that exist in the candidate samples but are unseen in the virtual training set.
Then for each selected composition, we further check if its individual words exist in the virtual training set.
If so, such a composition will be identified as a word-word level novel composition w.r.t. the virtual training set.
For example, if the composition ``white bird" is unseen in the virtual training set, while its individual words ``white" and ``bird" both exist in the virtual training set, this composition is a word-word level novel composition w.r.t. the virtual training set.
(iii) Finally, we select the candidate samples, of which the expressions contain the identified word-word level novel compositions w.r.t. the virtual training set, to construct the virtual testing set.

To efficiently perform the above steps, we can employ parallel matrix operations.
Specifically, to record the word-word level compositions in the virtual training set, we build a matrix  $\mathcal{M}_{v\_tr}$ with the shape of $|\mathcal{V}| \times |\mathcal{V}|$, where $|\mathcal{V}|$ is the size of the vocabulary set $\mathcal{V}$ of the original training set.
In this matrix, each element at the $i-th$ row and the $j-th$ column ($i,j \in\{1,\ldots,|\mathcal{V}|\}$) is a binary value, and the value 1 indicates that the $i-th$ word and $j-th$ word in the vocabulary set $\mathcal{V}$ form a word-word level composition in the virtual training set, while 0 means such a word-word level composition does not exist in the virtual training set.
Similarly, we can also build a matrix $\mathcal{M}_{candi}$ to record the word-word level compositions in the candidate samples.
Then to identify word-word level novel compositions w.r.t. the virtual training set ($\mathcal{M}_{v\_tr}$) from the candidate samples ($\mathcal{M}_{candi}$), we can efficiently compute a difference matrix: $\mathcal{M}_{diff} = \mathcal{M}_{candi} - \mathcal{M}_{v\_tr}$.
In $\mathcal{M}_{diff}$, any element with the value 1 indicates that the corresponding composition exists in the candidate samples, but is unseen in the virtual training set.
Then by further checking whether the individual words in such a composition exist in the virtual training set, we can determine if it is a word-word level novel composition w.r.t. the virtual training set.

Similarly, we can efficiently construct the other two virtual testing sets.
Note that to help the model to learn to handle a wide range of possible novel compositions,
at the beginning of each training epoch, we randomly sample a subset of the training data to re-construct the virtual training set, and leverage the remaining data to re-construct the virtual testing sets using the above strategy.

As discussed above, to handle various levels of novel compositions, we construct multiple virtual testing sets. A simpler alternative is to construct only one virtual testing set, which includes all the data samples containing any level of novel compositions w.r.t. the virtual training set.
However, compared to this alternative, by constructing multiple virtual testing sets, each level of novel compositions in the corresponding virtual testing set can offer an explicit generalization feedback during model training.
Thus the model is explicitly encouraged to well handle each level of novel compositions.
Moreover, constructing multiple virtual testing sets can facilitate the use of curriculum learning strategy for model training as discussed in Sec. \ref{sec.curriculum_training}.

\subsection{Training and Testing}\label{sec.curriculum_training}
To help models generalize to novel compositions in RES, 
our framework only changes the model training scheme. 
Thus our framework is general, and can be flexibly applied to train various RES models.
During training, at each epoch, we first split the training set to construct a virtual training set and multiple virtual testing sets.
Then we iteratively optimize the model over the virtual training set and virtual testing sets via meta optimization.
Specifically, for each meta optimization iteration, we use two batches of data samples: one batch of data for virtual training and the other batch of data for virtual testing.
Thus we need to ensure that the batch of data for virtual testing consists of the data samples containing novel compositions w.r.t. the batch of data for virtual training.
To this end, for each iteration, we randomly sample some data from each virtual testing set, to form the batch of data for virtual testing.
Then we select the virtual training set samples that contain all the individual components of the novel compositions in the batch of data for virtual testing, to form the batch of data for virtual training.
Note that the above data preparation procedures can be done before the model training.
For model testing, we test the trained model in the conventional way.

Besides, given that the various levels of novel compositions correspond to different levels of comprehension complexity, 
we adopt a \emph{curriculum learning} strategy \cite{bengio2009curriculum, wang2021survey} for model training, so that the model can progressively learn to handle various levels of novel compositions, i.e., from lower level (word-word) to middle level (word-phrase) to higher level (phrase-phrase), 
and thus can learn all these levels of novel compositions better.
Specifically, in the first $1/3$ of training epochs, we only use the virtual testing set for handling word-word level novel compositions for meta optimization.
Then in the middle $1/3-2/3$ of training epochs, we add the virtual testing set for handling word-phrase level novel compositions. 
Finally, in the remaining training epochs, we use all the three virtual testing sets.

\begin{table}[t]
   \caption{Comparison with the state-of-the-arts on three benchmark datasets using the metric of overall IoU.
      Moreover, we apply our framework on various models \cite{luo2020multi, yang2022lavt, zhu2022seqtr}, and obtain  consistent performance improvement. ``-'' indicates that the corresponding result is not provided in the original paper.
      U means the UMD split of RefCOCOg dataset, and G means the Google split.
   }
   \vspace{-0.35cm}
   \centering
   \begin{adjustbox}{max width=\linewidth}
      \setlength{\tabcolsep}{2pt}
      \begin{tabular}{lccccccccc} \hline
         \multirow{2}{4em}{Method}      & \multicolumn{3}{c}{RefCOCO} & \multicolumn{3}{c}{RefCOCO+} & \multicolumn{3}{c}{RefCOCOg}                                                                                                       \\ \cmidrule(lr){2-4} \cmidrule(lr){5-7} \cmidrule(lr){8-10}
                                        & val                         & testA                        & testB                        & val            & testA          & testB          & val(U)         & test (U)       & val (G)        \\ \hline\hline
         DMN \cite{margffoy2018dynamic} & 49.78                       & 54.83                        & 45.13                        & 38.88          & 44.22          & 32.29          & -              & -              & 36.76          \\
         RRN \cite{li2018referring}     & 55.33                       & 57.26                        & 53.93                        & 39.75          & 42.15          & 36.11          & -              & -              & 36.45          \\
         MAttNet \cite{yu2018mattnet}   & 56.51                       & 62.37                        & 51.70                        & 46.67          & 52.39          & 40.08          & 47.64          & 48.61          & -              \\
         CAC \cite{chen2019referring}   & 58.90                       & 61.77                        & 53.81                        & -              & -              & -              & 46.37          & 46.95          & 44.32          \\
         CMSA \cite{ye2019cross}        & 58.32                       & 60.61                        & 55.09                        & 43.76          & 47.60          & 37.89          & -              & -              & 39.98          \\
         STEP \cite{chen2019see}        & 60.04                       & 63.46                        & 57.97                        & 48.19          & 52.33          & 404.1          & -              & -              & 46.40          \\
         BRINet \cite{hu2020bi}         & 60.98                       & 62.99                        & 59.21                        & 48.17          & 52.32          & 42.11          & -              & -              & 48.04          \\
         CMPC \cite{huang2020referring} & 61.36                       & 64.53                        & 59.64                        & 49.56          & 53.44          & 43.23          & -              & -              & 49.05          \\
         LSCM \cite{hui2020linguistic}  & 61.47                       & 64.99                        & 59.55                        & 49.34          & 53.12          & 43.50          & -              & -              & 48.05          \\
         CMPC+ \cite{liu2021cross}      & 62.47                       & 65.08                        & 60.82                        & 50.25          & 54.04          & 43.47          & -              & -              & 49.89          \\
         EFN \cite{feng2021encoder}     & 62.76                       & 65.69                        & 59.67                        & 51.50          & 55.24          & 43.01          & -              & -              & 51.93          \\
         BUSNet \cite{yang2021bottom}   & 63.27                       & 66.41                        & 61.39                        & 51.76          & 56.87          & 44.13          & -              & -              & 50.56          \\
         CGAN \cite{luo2020cascade}     & 64.86                       & 68.04                        & 62.07                        & 51.03          & 55.51          & 44.06          & 51.01          & 51.69          & 46.54          \\
         LTS \cite{jing2021locate}      & 65.43                       & 67.76                        & 63.08                        & 54.21          & 58.32          & 48.02          & 54.40          & 54.25          & -              \\
         VLT \cite{ding2021vision}      & 65.65                       & 68.29                        & 62.73                        & 55.50          & 59.20          & 49.36          & 52.99          & 56.65          & 49.76          \\
         ReSTR \cite{kim2022restr}      & 67.22                       & 69.30                        & 64.45                        & 55.78          & 60.44          & 48.27          & -              & -              & 54.48          \\
         CRIS \cite{wang2022cris}       & 70.47                       & 73.18                        & 66.10                        & 62.27          & 68.08          & 53.68          & 59.87          & 60.36          & -              \\ \hline
         MCN \cite{luo2020multi}        & 62.44                       & 64.20                        & 59.71                        & 50.62          & 54.99          & 44.69          & 49.22          & 49.40          & -              \\
         MCN + Ours                     & \textbf{64.51}              & \textbf{66.48}               & \textbf{62.84}               & \textbf{52.29} & \textbf{56.71} & \textbf{46.72} & \textbf{51.49} & \textbf{51.63} & -              \\ \hline
         SeqTR \cite{zhu2022seqtr}      & 71.70                       & 73.31                        & 69.82                        & 63.04          & 66.73          & 58.97          & 64.69          & 65.74          & -              \\
         SeqTR + Ours                   & \textbf{73.23}              & \textbf{75.01}               & \textbf{71.95}               & \textbf{64.71} & \textbf{67.85} & \textbf{60.85} & \textbf{66.77} & \textbf{67.48} & -              \\ \hline
         LAVT \cite{yang2022lavt}       & 72.73                       & 75.82                        & 68.79                        & 62.14          & 68.38          & 55.10          & 61.24          & 62.09          & 60.50          \\
         LAVT + Ours                    & \textbf{74.92}              & \textbf{76.98}               & \textbf{70.84}               & \textbf{64.32} & \textbf{69.68} & \textbf{56.64} & \textbf{63.51} & \textbf{64.90} & \textbf{61.63} \\ \hline
      \end{tabular}
   \end{adjustbox}
   \vspace{-0.5cm}
   \label{Tab:RES_sota}
\end{table}
\begin{table}[h]
   \caption{We evaluate our framework on different testing subsets w.r.t. novel compositions to validate its effectiveness on optimizing the model to better generalize to novel compositions. We use the metric of overall IoU here.
   The performance gain of our framework compared to the corresponding baseline model is shown in parentheses.}
   \vspace{-0.4cm}
   \centering
   \begin{adjustbox}{max width=\linewidth}
      \setlength{\tabcolsep}{5pt}
      \begin{tabular}{lcccccccccccc} \hline
         \multirow{2}{4em}{Method} & \multicolumn{2}{c}{RefCOCO-val} & \multicolumn{2}{c}{RefCOCO-testA} & \multicolumn{2}{c}{RefCOCO-testB}                             \\ \cmidrule(lr){2-3} \cmidrule(lr){4-5} \cmidrule(lr){6-7}
                                   & Novel                       & Non-novel                             & Novel                         & Non-novel & Novel & Non-novel \\ \hline\hline
         MCN [23]   & 53.17 &  67.41 & 55.43  & 70.08 & 50.64 & 66.47 \\
         MCN + Ours                &  \textbf{57.38 ($\uparrow$4.21)} &  \textbf{68.09 ($\uparrow$0.68)} & \textbf{59.66 ($\uparrow$4.23)} & \textbf{70.67 ($\uparrow$0.59)} & \textbf{55.87 ($\uparrow$5.23)} & \textbf{67.04 ($\uparrow$0.57)} \\ \hline
         SeqTR [40] & 64.24 & 78.59 & 65.71  & 79.46  &  60.14 & 77.21\\
         SeqTR + Ours             & \textbf{67.31 ($\uparrow$3.07)} &  \textbf{79.06 ($\uparrow$0.47)} & \textbf{69.48 ($\uparrow$3.77)} & \textbf{79.84 ($\uparrow$0.38)} &  \textbf{64.89 ($\uparrow$4.75)} & \textbf{77.68 ($\uparrow$0.47)}\\ \hline
         LAVT [35]          &  63.52                         &  78.55                            & 67.17                              & 80.71 &  58.49    & 76.23  \\
         LAVT + Ours                      &  \textbf{67.42 ($\uparrow$3.90)}                         &  \textbf{79.29 ($\uparrow$0.74)}                          & \textbf{70.05 ($\uparrow$2.88)}                          &  \textbf{81.37 ($\uparrow$0.66)}    & \textbf{62.63 ($\uparrow$4.14)}      & \textbf{76.92 ($\uparrow$0.69)}  \\ \hline
      \end{tabular}
   \end{adjustbox}
   \vspace{-0.2cm}
   \label{Tab:ablation_1}
\end{table}

\section{Experiments}
We evaluate our method on three commonly used RES benchmarks: RefCOCO \cite{yu2016modeling}, RefCOCO+ \cite{yu2016modeling} and RefCOCOg \cite{mao2016generation, nagaraja2016modeling}.
Images in these three benchmarks are all from the COCO dataset \cite{lin2014microsoft}.
RefCOCO \cite{yu2016modeling} contains 19994 images, 50000 annotated objects with 142209 referring expressions.
RefCOCO+ \cite{yu2016modeling} consists of 19992 images with 49856 annotated objects, and 141564 expressions.
Different from RefCOCO, words describing absolute spatial locations (e.g., left, front) are not allowed to be used in the expressions in RefCOCO+.
In RefCOCO and RefCOCO+ datasets, following the original split in \cite{yu2016modeling}, 
the visual entities to be segmented in testA subset are people, while the ones in testB subset are objects (i.e., not people).
RefCOCOg \cite{mao2016generation, nagaraja2016modeling} includes 26711 images, 54822 annotated objects and 104560 expressions.
Compared with RefCOCO and RefCOCO+, the expressions in RefCOCOg are more complex, which have an average length of 8.4 words.
There exist two different partitions for RefCOCOg dataset: UMD split \cite{nagaraja2016modeling} and Google split \cite{mao2016generation}.

\textbf{Evaluation metrics.}
Following \cite{yang2022lavt, ye2019cross, chen2019see}, we report our results using two kinds of metrics: overall IoU (oIoU) and Precision@X (P@X).
The overall IoU measures the ratio of total intersection regions over total union regions of predicted masks and ground truths of all testing samples.
Precision@X calculates the percentage of testing samples, of which the model prediction has an IoU score higher than the threshold value X, and X$\in\{0.5, 0.6, 0.7, 0.8, 0.9\}$.

\textbf{Implementation details.}
We conduct our experiments on 8 Tesla V100 GPUs.
We applied our framework on various RES models \cite{zhu2022seqtr, yang2022lavt, luo2020multi}.
On each dataset,  we randomly sample $60\%$ of the training data as the virtual training set and use the remaining training data to construct virtual testing sets at each training epoch.
The learning rate ($\alpha$) for virtual training is 5e-5, and the learning rate ($\beta$) for meta update is 2e-5.

\subsection{Experimental Results}
As shown in Table \ref{Tab:RES_sota}, 
our framework achieves state-of-the-art performance across all three datasets, demonstrating the superiority of our framework.
Moreover, we applied our framework on two transformer-based SOTA models \cite{zhu2022seqtr, yang2022lavt} and a CNN-based model \cite{luo2020multi}. Our framework brings consistent performance improvement on all these models and datasets.
This shows that our framework can serve as a general approach to enhance model performance

\begin{figure}[t]
   \centering
   \includegraphics[width=0.95\linewidth]{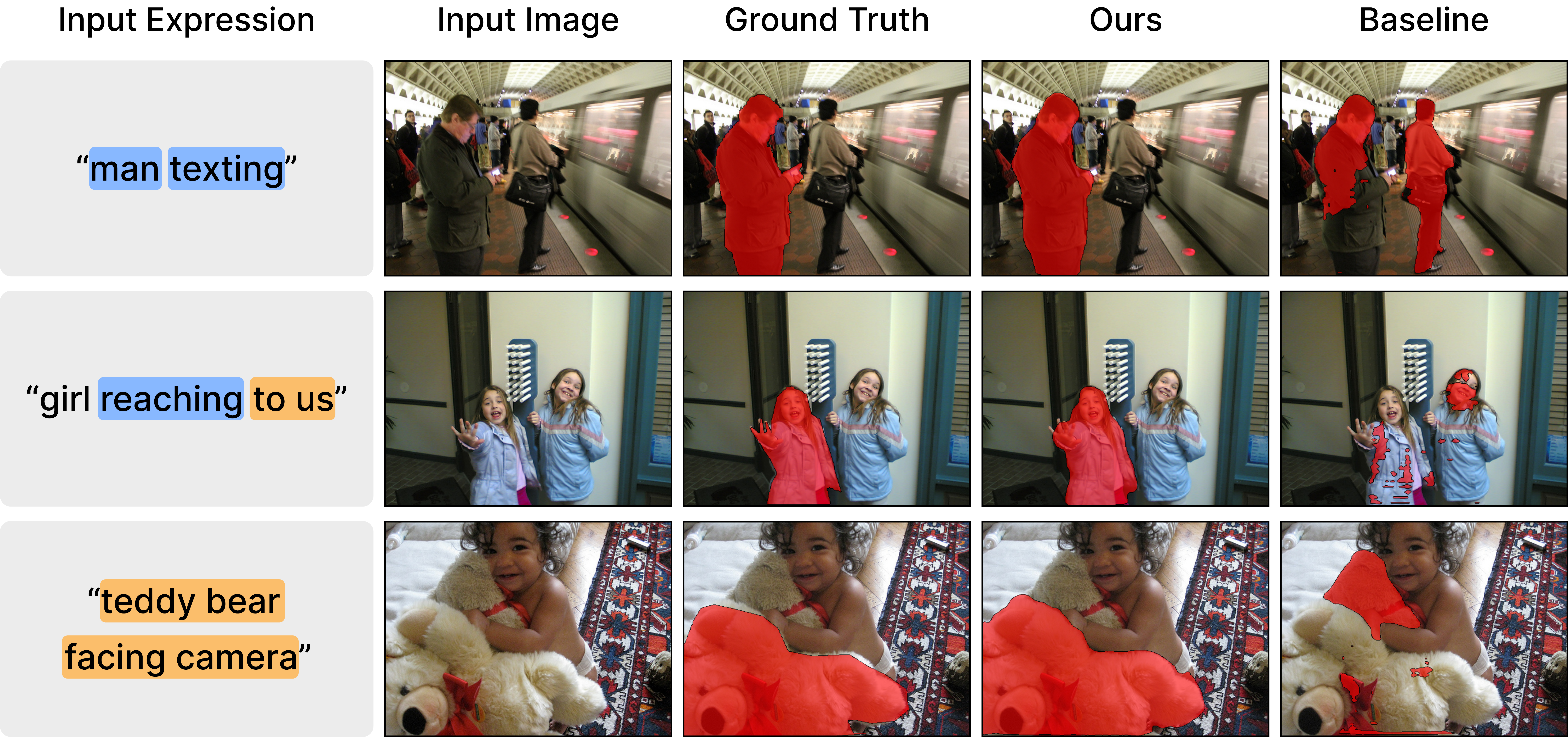}
   \vspace{-0.3cm}
   \caption{\textbf{Qualitative results of our method and the baseline model \cite{yang2022lavt}.} The above testing examples contain a word-word novel composition, a word-phrase novel composition, and a phrase-phrase novel composition respectively. As shown, by applying our framework on the baseline model, our method performs better when handling novel compositions of the learned concepts.}
   \label{fig:qualitative}
   \vspace{-0.6cm}
\end{figure}

To further validate the effectiveness of our framework for handling novel compositions, we perform the following analysis.
Specifically, we split each testing set of RefCOCO dataset to construct two subsets. 
One subset (\emph{Non-novel}) includes the data samples, in which all the contained compositions are seen in the RefCOCO training set.
While another subset (\emph{Novel}) includes the data samples that contain any level of novel compositions w.r.t. the training set.
As shown in Table \ref{Tab:ablation_1}, on each testing set, we can see a clear performance gap between the two subsets for each of the baseline models \cite{yang2022lavt, luo2020multi, zhu2022seqtr}, showing that existing models struggle with handling novel compositions.
Then by applying our framework on each baseline model, we obtain significant performance improvement on the subset of \textit{Novel}.
This validates the general effectiveness of our framework to optimize the model to well generalize to novel compositions.
Moreover, our framework also slightly improves the model performance on the subset of \textit{Non-novel}.
This can be attributed to that by training the model to better capture semantics and visual representations of individual concepts, our framework can help the model to better understand the given expression and find its visual correspondence, which thus can generally improve model performance.
We also show some qualitative results in Fig. \ref{fig:qualitative}.
As shown, when handling the testing samples containing novel compositions, our framework achieves better performance than the baseline model \cite{yang2022lavt}.

\subsection{Ablation Studies}
Following \cite{yang2022lavt, ye2019cross, chen2019see}, we conduct ablation experiments to evaluate our framework on RefCOCO validation set.
\begin{table}[t]
   \caption{We test several variants to investigate the impact of each level of novel compositions on model performance.}
   \vspace{-0.35cm}
   \centering
   \begin{adjustbox}{max width=\linewidth}
      \setlength{\tabcolsep}{2pt}
      \begin{tabular}{lcccccc} \hline
         Method                               & oIoU           & P@0.5          & P@0.6          & P@0.7          & P@0.8          & P@0.9          \\ \hline\hline
         Baseline (LAVT)                      & 72.73          & 84.46          & 81.24          & 75.28          & 64.71          & 34.30          \\ \hline
         w/o word-word novel compositions     & 73.43          & 84.71          & 81.69          & 76.12          & 65.23          & 34.47          \\
         w/o word-phrase novel compositions   & 73.68          & 84.82          & 81.76          & 76.28          & 65.37          & 34.56          \\
         w/o phrase-phrase novel compositions & 73.59          & 84.73          & 81.73          & 76.16          & 65.35          & 34.45          \\ \hline
         Ours                                 & \textbf{74.92} & \textbf{86.23} & \textbf{83.45} & \textbf{77.25} & \textbf{66.56} & \textbf{35.61} \\ \hline
      \end{tabular}
   \end{adjustbox}
   \vspace{-0.45cm}
   \label{Tab:ablation_2}
\end{table}
\begin{table}[t]
   \caption{We evaluate a variant to test the effectiveness of meta optimization scheme in our framework.
      For this variant, its optimization objective is to minimize $\mathcal{L}_{v\_tr}(\theta) + \sum_{k=1}^K \mathcal{L}_{v\_te}^k(\theta)$ (i.e., replacing $\theta^{\prime}$ with $\theta$ in Eqn. \ref{eq:optimization-objective}).}
   \vspace{-0.3cm}
   \centering
   \begin{adjustbox}{max width=\linewidth}
      \setlength{\tabcolsep}{6pt}
      \begin{tabular}{lcccccc} \hline
         Method            & oIoU           & P@0.5          & P@0.6          & P@0.7          & P@0.8          & P@0.9          \\ \hline\hline
         Training w/o meta & 72.76         & 84.50         & 81.27          & 75.33          & 64.72         & 34.31         \\ \hline
         Ours              & \textbf{74.92} & \textbf{86.23} & \textbf{83.45} & \textbf{77.25} & \textbf{66.56} & \textbf{35.61} \\ \hline
      \end{tabular}
   \end{adjustbox}
   \vspace{-0.6cm}
   \label{Tab:ablation_3}
\end{table}

\textbf{Impact of various levels of novel compositions.}
To investigate the impact of each level of novel compositions on model performance,
we test multiple variants.
One variant (\textit{w/o word-word novel compositions}) ignores the word-word level novel compositions.
This means that in this variant, we omit the virtual testing set for handling word-word level novel compositions during meta optimization.
Similarly, the other variants (\textit{w/o word-phrase novel compositions} and \textit{w/o phrase-phrase novel compositions}) ignore the corresponding level of novel compositions respectively.
As shown in Table \ref{Tab:ablation_2}, ignoring any level of novel compositions leads to performance drop compared to our framework, demonstrating that each level of novel compositions can affect model generalization performance.

\textbf{Impact of meta optimization.}
To investigate the effectiveness of the meta optimization scheme in our framework, we evaluate a variant (\textit{training w/o meta}).
In this variant, we train the model in the conventional manner on the constructed virtual training set and virtual testing sets, i.e., without meta optimization.
Note that for this variant, we still construct the virtual training set and virtual testing sets in the same way as in our framework.
As shown in Table \ref{Tab:ablation_3}, our framework outperforms this variant obviously, demonstrating the effectiveness of our meta optimization scheme.

\textbf{Impact of multiple virtual testing sets and curriculum learning.}
In our framework, we construct multiple virtual testing sets to handle various levels of novel compositions.
Moreover, since the various levels of novel compositions correspond to different levels of comprehension complexity, we adopt a curriculum learning strategy to facilitate model training.
To investigate the effectiveness of such design, we evaluate two variants.
One variant (\emph{one virtual testing set}) constructs only one virtual testing set to handle all levels of novel compositions. 
Such a virtual testing set includes all the data samples containing any level of novel compositions w.r.t. the virtual training set.
Another variant (\emph{multiple virtual testing sets w/o curriculum learning}) constructs multiple virtual testing sets as in our framework, but does not adopt the curriculum learning strategy. 
We compare these two variants to our original framework (\emph{multiple virtual testing sets w/ curriculum learning}).

As shown in Table \ref{Tab:ablation_4}, compared to the variant constructing one virtual testing set, the other variant 
obtains better performance, showing the superiority of using multiple virtual testing sets to handle various levels of novel compositions.
Furthermore, by employing the curriculum learning strategy, our framework performs better than the variant constructing multiple virtual testing sets without curriculum learning, demonstrating the effectiveness of our curriculum learning strategy.

\textbf{Impact of the size of virtual training set.}
In our framework, we randomly sample $60\%$ of the training data as the virtual training set, and use the remaining $40\%$ of the training data to construct virtual testing sets (\emph{60\%:40\%}).
Here we test two variants.
One variant (\emph{50\%:50\%}) uses $50\%$ of the training data to construct the virtual training set, and the remaining $50\%$ to construct virtual testing sets.
While another variant (\emph{70\%:30\%}) uses $70\%$ of the training data for the virtual training set, and the remaining $30\%$ part for virtual testing sets.
As shown in Table \ref{Tab:supportsetsize}, our method and these two variants all perform better than the baseline model (i.e., LAVT \cite{yang2022lavt}), showing the robustness of our framework in terms of the varying size of virtual training set.

\textbf{Impact of virtual testing sets construction strategy.}
In our framework, after sampling a subset of the training set as the virtual training set, we use the remaining training data to construct multiple virtual testing sets.
Each virtual testing set consists of the data samples containing one level of novel compositions w.r.t. the virtual training set.
To explore the efficacy of such a strategy, we evaluate a variant (\emph{random virtual testing sets}), in which  
we totally \emph{randomly} select training data to construct each virtual testing set.
As shown in Table \ref{Tab:querysets}, our framework obviously outperforms this variant.
This shows that our virtual testing sets construct strategy can effectively help our framework to improve model generalization performance.

\textbf{Training time.}
As shown in Table \ref{Tab:ablation_5}, we test the training time of our framework that trains the baseline network \cite{yang2022lavt} with meta optimization, and compare it to the training time of the baseline that trains the same network in the conventional manner without meta optimization, on RefCOCO dataset.
Though our framework performs better, it brings only relatively little increase ($18.18\%$) in training time.

\textbf{Impact of additional gradient updates.}
As discussed above, compared to the baseline model, our framework trains the model for longer time and involves more gradient updates. 
To explore whether the performance improvement of our framework comes from the additional gradient updates, we test a variant (\emph{baseline w/ additional gradient updates}) in which we train the baseline model (following the original training strategy) for as many iterations as in our framework. 
As shown in Table \ref{Tab:ablation_additional}, the performance of this variant is very close to the baseline model \cite{yang2022lavt}, and is obviously worse than our framework.
This might be because that the originally trained baseline models have already reached convergence under the original training strategy, and thus additional gradient updates would not bring obvious benefits.
Such results further validate the effectiveness of our framework.

\begin{table}[t]
   \caption{We evaluate two variants to test the impact of using multiple virtual testing sets and the curriculum learning strategy.}
   \vspace{-0.3cm}
   \centering
   \begin{adjustbox}{max width=\linewidth}
      \setlength{\tabcolsep}{2pt}
      \begin{tabular}{lcccccc} \hline
         Method            & oIoU           & P@0.5          & P@0.6          & P@0.7          & P@0.8          & P@0.9          \\ \hline\hline
         Ours (one virtual testing set)     &  73.64        & 84.82          & 82.08          & 76.24         & 65.47          & 34.64            \\ 
         Ours (multiple virtual testing sets w/o curriculum learning) & 74.02    &   85.34  & 82.63    &   76.87  & 66.09   &  34.95 \\ \hline
         Ours (multiple virtual testing sets w/ curriculum learning) & \textbf{74.92} & \textbf{86.23} & \textbf{83.45} & \textbf{77.25} & \textbf{66.56} & \textbf{35.61} \\ \hline
      \end{tabular}
   \end{adjustbox}
   \vspace{-0.40cm}
   \label{Tab:ablation_4}
\end{table}
\begin{table}[t]
   \caption{We test different variants that utilize different proportions of the training data to construct the virtual training set and virtual testing sets.}
   \vspace{-0.3cm}
   \centering
   \begin{adjustbox}{max width=\linewidth}
      \setlength{\tabcolsep}{4pt}
      \begin{tabular}{lcccccc} \hline
        Method                               & oIoU           & P@0.5          & P@0.6          & P@0.7          & P@0.8          & P@0.9          \\ \hline\hline
        Baseline (LAVT)                      & 72.73          & 84.46          & 81.24          & 75.28          & 64.71          & 34.30          \\ \hline
        Ours (50\%:50\%)                           & 74.78          & 86.11          & 83.32          & 77.08          & 66.38          & 35.54          \\
        Ours (60\%:40\%)                            & 74.92          & 86.23          & 83.45          & 77.25          & 66.56          & 35.61           \\ 
        Ours (70\%:30\%)                            & 74.74          & 86.04          & 83.26          & 77.01          & 66.29          & 35.52            \\ \hline
      \end{tabular}
   \end{adjustbox}
   \vspace{-0.40cm}
   \label{Tab:supportsetsize}
\end{table}
\begin{table}[t]
   \caption{We evaluate a variant to investigate the efficacy of virtual testing sets construction strategy in our framework.}
   \vspace{-0.3cm}
   \centering
   \begin{adjustbox}{max width=\linewidth}
      \setlength{\tabcolsep}{4pt}
      \begin{tabular}{lcccccc} \hline
         Method                               & oIoU           & P@0.5          & P@0.6          & P@0.7          & P@0.8          & P@0.9          \\ \hline\hline
         Random virtual testing sets                    & 72.81          & 84.56          &  81.28         &  75.34         & 64.74          & 34.32       \\ \hline
         Ours                                 & \textbf{74.92} & \textbf{86.23} & \textbf{83.45} & \textbf{77.25} & \textbf{66.56} & \textbf{35.61} \\ \hline
      \end{tabular}
   \end{adjustbox}
   \vspace{-0.4cm}
   \label{Tab:querysets}
\end{table}
\begin{table}[t]
   \caption{Comparison of the training time. Note that our method achieves much better performance than the baseline.}
   \vspace{-0.3cm}
   \centering
   \begin{adjustbox}{max width=\linewidth}
      \setlength{\tabcolsep}{8pt}
      \begin{tabular}{lccccccc} \hline
         Method            &    Training time    & oIoU           & P@0.5          & P@0.6          & P@0.7          & P@0.8          & P@0.9          \\ \hline\hline
         Baseline          &       33 hours           & 72.73          & 84.46          & 81.24          & 75.28          & 64.71          & 34.30   \\ \hline
         Ours              &       39 hours           & 74.92         & 86.23           & 83.45          & 77.25          & 66.56          & 35.61 \\ \hline
      \end{tabular}
   \end{adjustbox}
   \vspace{-0.4cm}
   \label{Tab:ablation_5}
\end{table}
\begin{table}[t]
   \caption{We test a variant to investigate the impact of additional gradient updates on model performance.}
   \vspace{-0.3cm}
   \centering
   \begin{adjustbox}{max width=\linewidth}
      \setlength{\tabcolsep}{2pt}
      \begin{tabular}{lcccccc} \hline
         Method                               & oIoU           & P@0.5          & P@0.6          & P@0.7          & P@0.8          & P@0.9          \\ \hline\hline
         Baseline                                & 72.73          & 84.46          & 81.24          & 75.28          & 64.71          & 34.30          \\ \hline
         Baseline w/ additional gradient updates   & 72.74          & 84.43          & 81.25          & 75.25          & 64.73          & 34.28           \\ \hline
         Baseline w/ ours                          & \textbf{74.92} & \textbf{86.23} & \textbf{83.45} & \textbf{77.25} & \textbf{66.56} & \textbf{35.61} \\ \hline
      \end{tabular}
   \end{adjustbox}
   \vspace{-0.4cm}
   \label{Tab:ablation_additional}
\end{table}

\section{Conclusion}
In this work, we proposed a meta learning-based framework (MCRES) to improve the generalization performance of RES models, especially when handling novel compositions of learned concepts.
By constructing a virtual training set and multiple virtual testing sets w.r.t. various levels of novel compositions and then optimizing the model via meta optimization, our framework can effectively improve model generalization performance.
Extensive experiments show that our framework achieves superior performance on widely used benchmarks.
Moreover, our framework is flexible, and can be seamlessly applied on various models with different architectures to enhance their performance.

\noindent{\bf Acknowledgement.} This work is supported by MOE AcRF Tier 2 (Proposal ID: T2EP20222-0035), National Research Foundation Singapore under its AI Singapore Programme (AISG-100E-2020-065), and SUTD SKI Project (SKI 2021\_02\_06).

   {\small
      \bibliographystyle{ieee_fullname}
      \bibliography{main}
   }

\end{document}